# Marine Debris Detection in Satellite Surveillance using Attention Mechanisms

Ao Shen, Yijie Zhu and Richard Jiang

*Abstract*—Marine debris is an important issue for environmental protection, but current methods for locating marine debris are yet limited. In order to achieve higher efficiency and wider applicability in the localization of Marine debris, this study tries to combine the instance segmentation of YOLOv7 with different attention mechanisms and explores the best model. By utilizing a labelled dataset consisting of satellite images containing ocean debris, we examined three attentional models including lightweight coordinate attention, CBAM (combining spatial and channel focus), and bottleneck transformer (based on self-attention). Box detection assessment revealed that CBAM achieved the best outcome (F1 score of 77%) compared to coordinate attention (F1 score of 71%) and YOLOv7/bottleneck transformer (both F1 scores around 66%). Mask evaluation showed CBAM again leading with an F1 score of 73%, whereas coordinate attention and YOLOv7 had comparable performances (around F1 score of 68%/69%) and bottleneck transformer lagged behind at F1 score of 56%. These findings suggest that CBAM offers optimal suitability for detecting marine debris. However, it should be noted that the bottleneck transformer detected some areas missed by manual annotation and displayed better mask precision for larger debris pieces, signifying potentially superior practical performance.

## I. INTRODUCTION

MARINE debris is any persistent solid material that has been abandoned directly or indirectly, intentionally or unintentionally, into the Marine environment or the Great Lakes. Anything man-made and solid that is lost in these aquatic environments becomes Marine debris [1]. According to Maes et al. (2021), the main source of Marine debris is plastic [2]. The slow degradation and light weight of plastic make it easier to stay and accumulate in nature than other garbage, and even form small "islands" in the ocean that can be easily seen and intercepted. But more often, it's a garbage patch -- an area of the ocean made almost entirely of tiny plastic that's not always visible to the naked eye. It looks like cloudy soup mixed with things like fishing gear and shoes, such as the famous Great Pacific Garbage Patch (GPGP). Within this 1.6 million km2 area, there are 42,000 metric tons of giant plastics, 20,000 metric tons of large plastics, 10,000 metric tons of medium plastics and 6,400 metric tons of microplastics [3].

In addition to the pollution to the appearance of water bodies mentioned above, the most serious aspect of Marine debris is the harm to the Marine ecosphere. In the case of sea birds, the livers of black-footed albatrosses from Midway Atoll, near GPGP, were found to have high levels of perfluoroalkyl acid (PFAA), much higher than those found in waterbirds of the continental United States and Arctic environments. The concentration of PFOS ranged from 22.91 to 70.48 ng/g wet weight, and PFUdA ranged from 8.04 to 18.70 ng/g wet weight, but the concentration of the latter was much higher than that of the former in contaminated albatross livers [4]. And that's because of the plastic waste in GPGP.

In addition, Marine plastic has an impact on our daily lives as it enters the food chain. Marine plastic debris has been discovered in beer, salt, drinking water, vegetable-growing soil, and salt. Developmental, neurological, reproductive, and immunological issues can result from plastic materials' endocrine system disruption and carcinogenicity [5]. Toxic contaminants that frequently build up on plastic surfaces and are then ingested by people through seafood intake are another health risk.

At present, the mainstream detection method for marine debris is still traditional manual methods. However, conducting on-site investigations requires extensive labour, equipment, and financial expenditures, thereby limiting the scope. With the improvement of machine learning, especially image segmentation technology [6], the plastic fragments in the visual data captured by the camera system are gradually accepted by researchers. These visual data are mainly collected from airplanes and drones. However, the data acquisition range and efficiency of aerial photography are limited, especially for the lack of detection capabilities for distant sea targets. Therefore, detection using satellite imagery is an effective option. At present, there are not many studies using satellite imagery to detect marine debris, the scope is limited and the image segmentation method used is not the most novel method.

While there exist several conventional image segmentation algorithms for satellite images, these do not specifically target the ocean surface with sufficient detail due to limitations in resolution. Meanwhile, utilizing more sophisticated models could also help increase overall performance, making them useful in practical applications. One potential option would be adopting the use of YOLO architecture, which could provide greater speed and the ability for real-time processing

The manuscript was submitted on xx/xx/2023. This work was supported in part by the Engineering and Physical Sciences Research Council (EPSRC) under Grant EP/P009727/1, and the Leverhulme Trust under Grant RF-2019-492. (Corresponding author: Richard Jiang.)

Ao Shen, Yijie Zhu and Richard Jiang are with LIRA Center, Lancaster University, Lancaster, LA1 4YW, United Kingdom (E-mail: r.jiang2@lancaster.ac.uk).

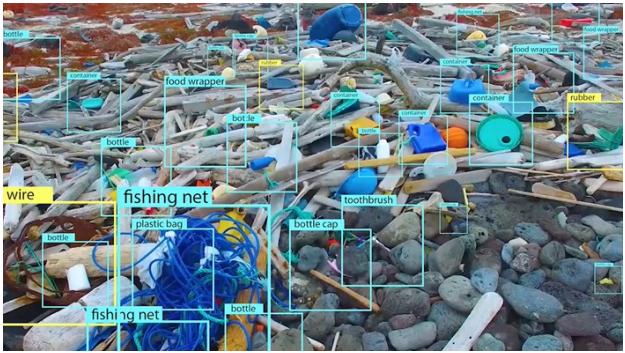

**Fig. 1.** An example of an object detection model applied to drone imagery, not a real application. This shows that the AI was able to identify the litter scattered on the beach.

capabilities, opening up new opportunities in remote sensing technologies.

Our research proposal aims to address the challenge of detecting marine debris through the combination of state-of-the-art technology and innovative methods. We plan to leverage the power of the neural network model based on YOLOv7 instance segmentation to achieve accurate identification and classification of marine debris in satellite imagery. By combining satellite technology and machine learning, the detection efficiency and cost of marine debris can be greatly improved. To further enhance the performance of our system, we intend to explore various attention mechanisms and evaluate their suitability for improving the results obtained from our baseline approach. By doing so, we hope to mitigate the shortcomings associated with R-CNN-based methods and ultimately yield improved detection rates and spatial precision. Ultimately, our goal is to develop an effective solution for monitoring and managing the growing problem of marine pollution.

## II. Related Work

### A. Current Methods for Marine Debris Detection

Due to the increasing importance of environmental protection, there has been a lot of research on the source and location of Marine debris, or Marine plastic. As one might expect, a number of traditional, manual methods have been used to count this Marine debris.

NOAA (National Oceanic and Atmospheric Administration) has a Marine Debris Monitoring and Assessment program, or MDMAP. The work works by recruiting partners and volunteers around the world to conduct regular surveys of plastic waste along coastlines. Monitoring kits will be made available for partners and volunteers to use uniform standards and methods to collect and record the amount and type of plastic waste. In this way, the quality and comparability of data can be guaranteed. By analyzing the survey data, they were able to understand the scale and trends of the problem, as well as the most common types of plastic waste in different regions. Since its inception in 2012, the work has conducted 4,421 surveys at 335 monitoring sites in nine countries [8]. They found that, globally, the most common pieces of plastic litter on coastlines were cigarette butts, food packaging, drink bottle caps and plastic bags. And the amount and type of plastic on the shoreline are related to population density, wind direction and tide. In addition, they analyzed the data and found that plastic waste from Asia reaches Hawaii's coastline on ocean currents. In the Caribbean, plastic waste on the coastline is mainly generated by local residents and tourists, rather than by ocean currents [9].

The NOAA's study highlights the benefits of in situ monitoring in measuring and analyzing the quantity and quality of plastic waste present in oceans and seas. On site measurements allow for precise evaluation of the geographical and chronological distribution of litter, thereby shedding light on the behavior patterns and effects of ocean currents. Additionally, through on-site observations, it becomes feasible to investigate the harmful consequences of plastic waste on aquatic species and habitats, facilitating risk assessments. Nonetheless, as mentioned previously, conducting fieldwork entails substantial expenditure of labor force, equipment, and funds, thus limiting coverage. Furthermore, even when tools are accessible, variations in proficiency and motivation among participants may result in unequal protocols and data inconsistencies. Therefore, alternative solutions must continue to be explored and developed.

With recent advancements in technology, a number of machine-learning approaches that rely on convolutional neural networks have emerged for identifying plastic debris in visual data captured using camera systems, including those mounted on drones, satellites, and vessels. These methods share the characteristic of processing images or streaming media content to locate plastic items within the recorded footage. Although primarily focused on aerial and spaceborne imagery, along with shipboard observations, these studies have shown promising results in addressing the issue of marine plastic pollution via computer vision techniques.

Typically, drones or aerial photography are used to observe Marine debris deposition along the coast. Ellipsis Earth uses drones equipped with cameras to map the location of plastic pollution. By applying image recognition based on deep learning, it can identify the type and the size of plastic. In some cases, even the brand or source of the waste can be identified. The drone can take video at high speeds in locations where humans can't go, and then fly back to read the data in its storage and feed it into an AI to identify plastic waste in it. Some more advanced algorithms have been applied to support real-time recognition of images transmitted synchronously by the UAV, as shown in Figure 1. According to research by Song et al., conducted in 2022, the drone mapping method can be more accurate and universal than traditional methods [10]. The segmentation model based on U-Net achieved good segmentation results in UAV image analysis, and the F1 scores of polypropylene foam and plastic, the most serious pollutants, were 0.97 and 0.93, respectively. However, there are limits to what the Ellipsis technique and UAV can detect. The drone approach is limited to coastal detection, not deep into the ocean. And it was unable to identify microplastics -- plastic particles smaller than 5 millimeters, of which at least 14 million tons are estimated to be on the ocean floor [11].

Another way to target large plastic waste is to use on-board cameras. This approach is an alternative to the traditional water collection detection method such as the neuston trawls. It uses an algorithm based on deep learning, which can detect plastic fragments from high-resolution optical image data collected from GoPro, and add a GPS tag to the image to quantify the plastic debris floating on the surface of the surface [12]. Since this method has a closer view than a drone, it can go deeper to provide more precise and detailed estimates of plastic density in the ocean.

Obviously, the above methods are mostly suitable for a small range of detection, and have the characteristics of long time consuming. To detect large amounts of microplastics and track the location of plastic debris at a macro level, satellites are required. A new method developed by researchers at the University of Michigan (UM) uses satellite data to map the concentrations of microplastics in the world's oceans [13]. Eight microsatellites that are a part of NASA's Cyclone Global Navigation Satellite System (CYGNSS) mission provided data to the researchers. In order to gauge the roughness of the ocean surface, the CYGNSS satellite collects signals reflected from Global Positioning System (GPS) satellites from the water. Waves are suppressed by plastic or other garbage in the ocean, causing a smaller roughness than anticipated. In a 2021 study, Jamali et al. used multispectral satellites and deep learning to develop a large-scale Marine plastic waste detection framework [14]. By combining Sentinel-2 satellite imagery with cutting-edge machine learning algorithms on the Sentinel Hub cloud API, they created a cloud-based framework for large-scale Marine pollution monitoring (API). Although both satellite-level plastic detection methods have achieved good results, the former is not universal and lacks detail, and the latter is still in the coastal part, which is not useful for deeper into the ocean.

Among current approaches for detecting marine debris, methods employing UAVs, vessel-mounted cameras, and satellite imagery possess varying strengths and weaknesses. While UAVs and ship-based cameras provide highly accurate and detailed information, their limited detection ranges and costs restrict their effectiveness in monitoring large regions. Satellite imaging, conversely, offers greater coverage but suffers from lower resolution and dearth of available data. Existing methods utilized for pinpointing plastic particles in satellite images mainly concentrate on coastal zones, leaving unexploited the unique capabilities of satellite imagery. As such, evaluating the performance of YOLO on high-resolution, surface-oriented satellite photographs represents a crucial step toward establishing a comprehensive system capable of detecting marine plastic debris over extensive territories while capitalizing on the unique advantages of satellite imagery.

*B. YOLO*

YOLO is a CNN-based architecture similar to a fully convolutional neural network. CNN is an artificial neural network similar to Deep Neural Networks (DNN). The inputs to the nodes in the neural network extract local features from each local corresponding field of the previous layer. After this feature extraction is completed, the positional relationships between local features and other features are mapped or plotted [15].

When the YOLO algorithm applies CNN, each convolutional network starts predicting multiple bounding boxes and bounding box class probabilities simultaneously. This means that instead of processing the image multiple times to detect different classes, YOLO simply passes the image through the neural network once to obtain a prediction or output. This optimizes the detection performance of a single algorithm run, thus reducing the latency of the process. This allows YOLO to identify and localize objects in near real-time, such as detecting objects in streaming videos.

As the first YOLO model with a new model head, YOLOv7 has both object detection and instance segmentation. The main advantage of YOLO is its speed. It can process images at 155 frames per second, which is much faster than other state-of-the-art algorithms [7]. Therefore, although its base implementation is Mask R-CNN, it is more efficient than traditional Mask R-CNN-based models and more suitable for real-time tasks.

In the recent COVID-19 pandemic, YOLO's efficiency makes it stand out and achieve excellent results in real-time mask detection tasks. In 2021, R. Liu and Z. Ren conducted a study to replace the traditional Faster R-CNN with YOLOv3 and YOLO achieves a higher F1 score and is nearly 50% Faster than Faster R-CNN [16].

YOLO also has a small number of applications in Marine debris detection, mainly its object detection function in the application of UAV aerial photography. A study by Jun-Ichiro Watanabe et al. used YOLOv3 as a deep learning object detection algorithm to detect debris floating on the beaches and the sea surface next to the beach. Through the detection of hand-taken plastic waste photos, YOLOv3 obtains 77.2% of the mAP, which is obviously higher precision and Faster than the 41.2% of Faster R-CNN [17]. Although the research is focused on in-atmosphere detection methods with drones, it confirmed the feasibility of the YOLO algorithm for Marine debris detection, and more importantly, it confirmed the excellent accuracy and identification speed of YOLO.

For the instance segmentation of YOLO, there are still relatively few practical applications, more are improved versions of YOLOv3 era, such as Poly-YOLO or Insta-YOLO [18][19]. These YOLO instance segmentation versions are based on backward YOLO versions and lack evaluation for real-world application scenarios. This highlights the importance of testing the application for YOLOv7 instance segmentation.

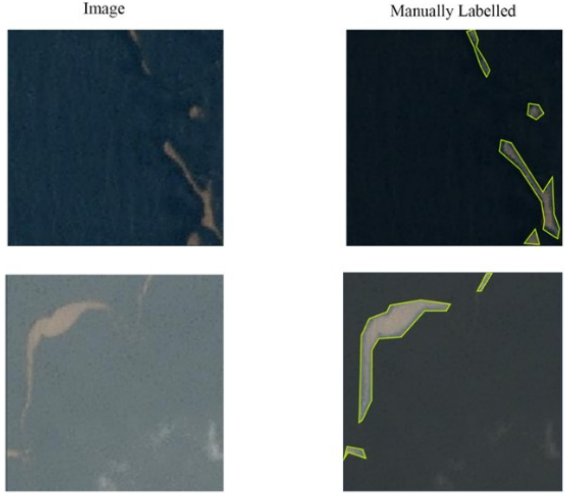

**Fig. 2.** Manually label process. On the left is the selected image, and on the right is the manually annotated image of the polygon suitable for instance segmentation.

### III. METHODS

*A. Data Collection and Labeling*

According to previous work, satellite images of Sentinel-2 have a resolution of 10 m per pixel, which is still not sufficient for accurate debris detection. For this work, NASA satellite images with a resolution of 3 meters per pixel from PlanetScope were used [20]. The pixel size of these images is 256x256·, with four bands of red, green, blue and near-infrared. The images are mostly of the Gulf Islands in Honduras, where the trash is made up of plastic, wood, algae, and other man-made objects.

As shown in Figure 2, although there are 707 images in this set, most of them have unclear garbage, lots of clouds, and coasts. For this reason, a total of 321 images were manually selected and labeled in two rounds using Roboflow (see Figure 2). Among them, 249 images were randomly selected to form the training set (79%) and 68 images were randomly selected to form the test set (21%) to form the data set used in this work.

*B. YOLOv7 Instance Segmentation*

Instance segmentation works by using a Convolutional neural network (CNN) to detect and classify each object in an image. Methods for instance segmentation can be divided into two categories. One of them can be called a two-stage method, which uses a Region Proposal Network (RPN) to generate candidate boxes in first, and then classifies and segments each box. Single-stage methods, on the other hand, perform classification and segmentation directly on the whole image without generating candidate boxes.

YOLOv7 is a single-stage instance method, which uses a novel Overlapping double structure (Overlapping BiLayers), is able to handle cover problem, and improve accuracy and speed. The network of YOLOv7 mainly includes four parts: Input, Backbone, Neck, and Head. Firstly, the image was preprocessed by a series of operations such as data enhancement in the input part, and then sent to the backbone, and the backbone extracted the features of the processed image. Then, the extracted features were fused by the neck module to obtain features of large, medium and small sizes. Finally, the fused features are sent to the head, and the results are output after detection.

According to C. Wang et al., the main backbone is mainly composed of convolution, E-ELAN module, MPConv module and SPPCSPC module [21]. ELAN is modified to E-ELAN in YOLOv7. Group convolutions are used by E-ELAN to increase the channels and cardinalities of the computation blocks. All computing blocks in the computation layer have the same channel multiplier and group parameters applied to them. Each computing block's feature maps are concatenated after being divided into groups of size g. To complete the combined cardinality, a shuffled group feature map will be added. The feature vectors are mapped to the number of classes here, which is two in this work (debris and no debris).

The Neck part of YOLOv7 consists of FPN feature pyramid and PANet. This part is the same as YOLOv5. The PANet part finally maps mask and box separately. Since this work is the detection of plastic debris, only two classes exist, with debris and without debris, so only "0" and "1" will be output when mapping the feature layer of the box.

This work uses a YOLOv7 instance segmentation version based on OpenCV and PyTorch [22]. A data control group was created by training the dataset directly on that version.

*C. YOLOv7 Instance Segmentation with C2f*

Due to the release of YOLOv8, this work also attempts to incorporate a part of YOLOv8 C2f module. As shown in Figure 3, the C2f module is a lightweight convolutional structure, which refers to the ideas of the C3 module and ELAN, and can obtain more abundant gradient flow information while ensuring lightweight. By replacing the specific convolution layer in YOLOv7, the C2f module is implanted without changing the overall structure of YOLOv7 to simulate a more lightweight scene.

As a variant of the basic YOLOv7, this version will be used as a base environment to test the effects of the attention mechanism. All models based on YOLOv7 with C2f added will be named "xx models with C2f" (where xx refers to an attention mechanism).

*D. Attention on YOLOv7 Segmentation*

Attention is a technique designed to mimic cognitive attention. It comes from the way humans perceive information about the environment. This effect enhances some parts of the input data while reducing others -- the motivation is that the network should pay more attention to smaller but important parts of the data.

*E. Coordinate Attention Application*

As a new mechanism based on channel attention, Hou et al. proposed a new attention mechanism, called coordinate attention, by embedding location information. Contrary to channel attention, which splits up feature tensors into a single feature vector by 2D global pooling, coordinate attention splits up channel attention into two 1D feature encoding procedures, each of which aggregates features along two different spatial directions [23]. This allows for the acquisition of remote dependencies along one spatial direction and the retention of accurate position data along the other. The direction-aware and location-sensitive attention maps that are produced from the resulting feature maps may then be complementary applied to the input feature map to improve the representation of the object of interest. Figure 4 shows the structure of a coordinate attention block.

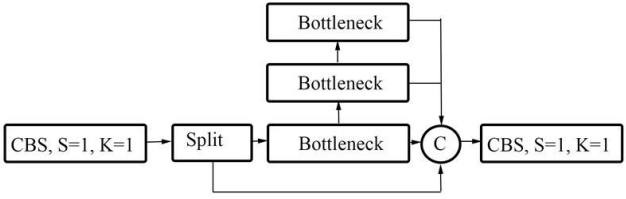

**Fig. 3.** Architecture diagram of C2f. Compared with the C3 module used in YOLOv7, it is more lightweight and has richer gradient information

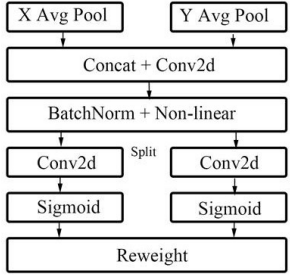

**Fig. 4.** Schematic expression of coordinate attention

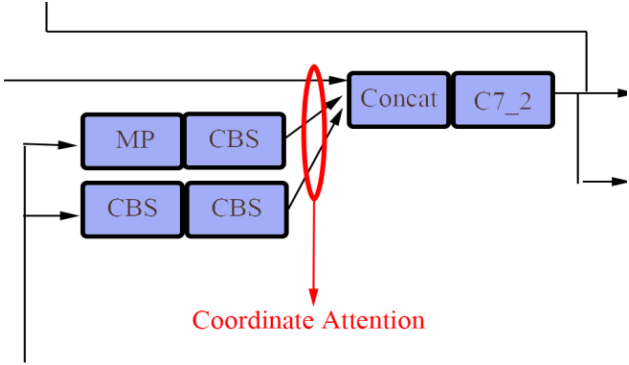

**Fig. 5.** Coordinate Attention model is created by insert coordinate attention blocks into the head of instance segmentation.

For an input X, as an example of coordinate information embedding, two spatial extents of pooling kernels (H,1) or (1, W) are carried to encode each channel through both horizontal and vertical positions. The output z of channel c at height h should be written as

$$z_c^h(h) = \frac{1}{W} \sum_{0 \le i < W} x_c(h, i) \quad (1)$$

Thus, the output of channel at width $w$ can be similarly formulated as

$$z_c^w(w) = \frac{1}{H} \sum_{0 \le j < H} x_c(j, w) \quad (2)$$

In this way, coordinate attention aggregates features along both spatial directions, yielding direction-aware feature maps in pair. Therefore, in this work, the coordinate attention mechanism will be loaded as an additional module and replace a certain convolutional layer after the MP module of the YOLOv7 head (see Figure 5). This type of improved structure is called Coordinate Attention model in the following chapter in this report, which is the first improvement. Coordinate Attention model is also established on a version of YOLOv7 instance segmentation which includes the addition of C2f

*F. Convolutional Block Attention Module Application*

Convolutional Block Attention Module (CBAM) stands for attention mechanism module of Convolutional module, which is an attention mechanism module that combines spatial and channel (See Figure 6).

Compared with senet's attention mechanism which only focuses on channels, it can achieve better results. Figure 7 shows the general structure of CBAM. As observed, the convolutional layer's output result will first proceed to the channel attention module to obtain a weighted result, then pass through the spatial attention module to obtain the final result by weighting. [24].

The Channel Attention Module compresses the feature map in the spatial dimension to obtain a one-dimensional vector before operation. Not only the Average pooling but also the Max pooling are taken into account when compressing in the spatial dimension. When conducting the gradient backpropagation computation, Average pooling provides feedback to every pixel on the feature map, but Maximum pooling only provides feedback to the gradient where the response is the biggest in the feature map. The mechanism can be expressed as follows:

$$M_c(F) = \sigma\big(MLP(AvgPool(F)) + MLP(MaxPool(F))\big) \quad (3)$$

Where MLP is a multi-layer perceptron, F is two different spatial context descriptors, and σ is a sigmoid function.

The feature map produced by the Channel Attention module serves as the input feature map for the Spatial Attention module. The Spatial Attention module compresses the channel and performs Average Pooling and Max Pooling in the channel dimension respectively. The Max Pooling operation extracts the maximum value from the channel by multiplying the height by the width. The Average Pooling operation simply extracts the average of the channels, again multiplying the height by the width. Then all the feature maps (all with 1 channel) are combined to produce a 2-channel feature map. The mechanism can be summarized as follows:

$$M_s(F) = \sigma\big(f^{7\times 7}([AvgPool(F); MaxPool(F)])\big) \quad (4)$$

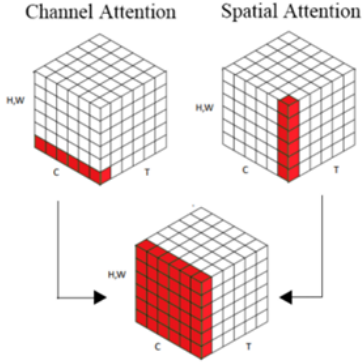

**Fig. 6.** What the combination of channel and spatial attention looks like. Where C represents the channel domain, H and W represent the spatial domain, and T represents the temporal domain.

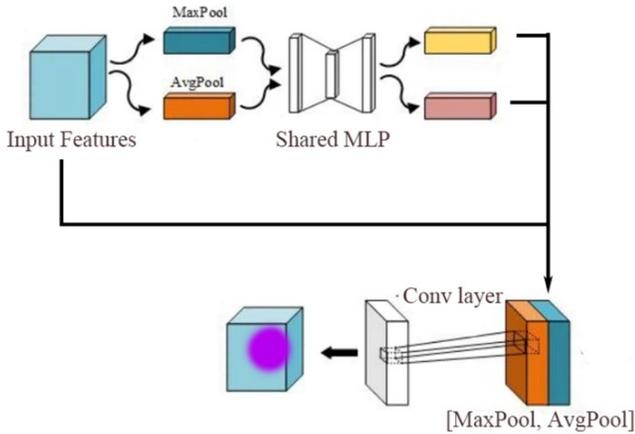

**Fig. 7.** Brief structure diagram of CBAM. The channel submodule uses the max-pooling output and the average pooling output of the shared network, and then obtains the weight of one channel submodule. The features weighted by the channel module will go to the spatial sub-module, which takes two similar outputs, pooling them along the channel axis and forwarding them to the convolutional layer.

Where $F$ is feature maps aggregated from channel information, σ is the sigmoid function and f is a convolution operation that has a filter size of 7x7.

By combining two attention modules in order, they can be complemented. In this way, the whole attention block is able to focus on both location and content. Among them, channel attention needs to be arranged before spatial attention. Thus, a new layer combined with CBAM block and original convolutional layer will be added in this work, replacing two original convolutional layers in the YOLOv7 backbone (as shown in Figure 8). A type of model will be formed in this way, which will be called CBAM model in the rest of this paper. A version of the model based on C2f is also created for comparison.

*G. Self-attention Application*

The self-attention mechanism or transformer is a variant of the attention mechanism that lessens reliance on outside input and excels at detecting internal correlations of features or data. The self-attention mechanism has been previously applied in text to solve the long-distance dependence problem by calculating the interaction between words. The core part of a self-attention mechanism in text processing is three vectors: Query, Key and Value. For a single word, Query is dot multiplied by the Key to get the score for the word, then a softmax calculation will provide the relevance of each word with respect to the current word. By adding the product of Value and softmax, value of self-attention at the current position can be obtained.

When used in computer vision, self-attention is often replaced with a convolutional layer [25]. It works similarly to text processing. For a pixel input $x_{ij}$, a local area in positions $ab \in N_k(i,j)$ is extracted, where $k$ is spatial extent centered around $x_{ij}$. Then a singled-headed self-attention with pixel output $y_{ij}$ can be roughly written as follows:

$$y_{ij} = \sum_{a,b \in N_k(i,j)} softmax_{ab}(q_{ij} k_{ab}) v_{ab} \quad (5)$$

Where $q_{ij}$, $k_{ab}$, $v_{ab}$ are queries, keys and values relatively. They provide linear transformations of pixels in position $ij$ and nearby. $softmax_{ab}$ illustrates a softmax based on computing all the logits in the neighborhood of $ij$.

Compared with the convolutional layer, which is a local operation and can only focus on a small region in the input feature map, the self-attention layer is a global operation and can focus on any position in the input feature map.

This work uses one of these Transformers: Bottleneck Transformer, which, similar to YOLO itself, is efficient and low-overhead. This mechanism is derived from the BoTNet architecture. The BoTNet design is simple: replace the last three spatial (3 × 3) convolutions in the ResNet with a Multi-Head Self-attention (MHSA) layer [26]. The core part of this architecture, MHSA, performs all2all attention on a 2D feature map (see Figure 9). It performs with split relative position encoding $R_h$ and $R_w$ for height and width, respectively. The attention logits are $qk^T$ and $qr^T$ in the diagram, where $q$, $k$, $r$ stands for query, key, and position encoding, respectively.

The self-attention used in the Bottleneck network is a transformer block within backbone. However, in order to facilitate the combination with other attention mechanisms, it is modified here to apply to the head part of the network. A new structure is created by inserting three segments of self-attention in the head, and a model is obtained through training. It will be called Self-attention model. A version based on the C2f module was also created since comparison. Figure 10 shows where the self-attention blocks are inserted.

*H. CBAM Self-attention*

In the preliminary training, CBAM model shows good performance, and Self-attention model shows good plasticity.

Therefore, an ensemble attempt was made to combine the two into a better structure. This design is to simultaneously consider the ability of the two attention mechanisms in the adoption of external information and internal information, making them complementary. The input feature map will first go to CBAM to pass the attention block based on external information, and then return to Self-attention to pass the attention block based on internal information. In order to test likelihood and maintain consistency, a layer combining CBAM block and convolutional layer is inserted into the CBS module after SPPCSPC module, see Figure 10.

This tentative architecture will be referred to as Dual-attention model in this paper. There is also a C2f module-based version of this structure for comparison.

IV. RESULTS

*A. Experimental Setup*

Each improved model, including YOLOv7 and YOLOv8 itself, is implemented in the Google Colab Python environment. The default number of epochs for YOLOv7 is 299; however, through several training runs, it is found that the last 100 epochs hardly change. So all models are trained for 199 epochs with a batch size of 4 to obtain the highest efficiency within the video memory limit. In addition, hyp.scratch-high is used as the hyperparameters setting. Finally, considering that lightweight and low cost is one of the performance indicators, it is necessary to ensure that other training parameters are consistent to observe the training time. The CBAM model has twice the training time of the other models, about 4 hours.

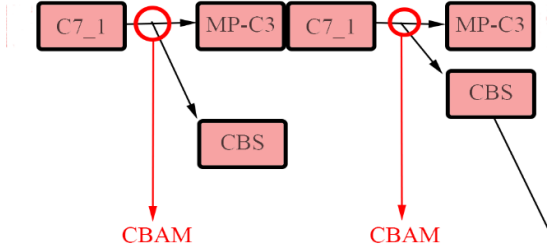

**Fig. 8.** CBAM model is created by replacing Conv layer in CBS modules with CBAM block. These two CBS modules are exits to branches in the backbone. A layer that is inserted before SPPCSPC module is tested, but has a poor result.

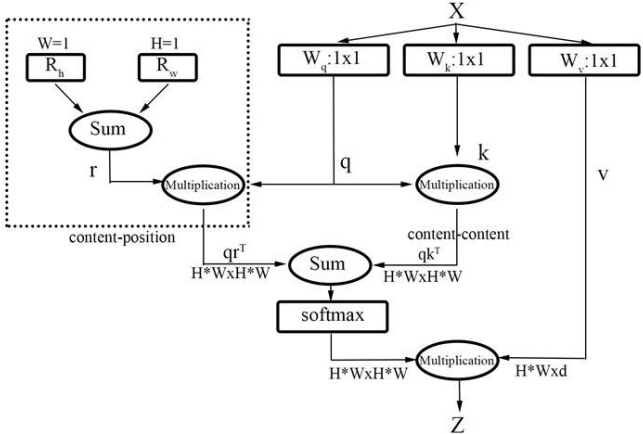

**Fig. 9.** Self-attention block used in BoTNet.

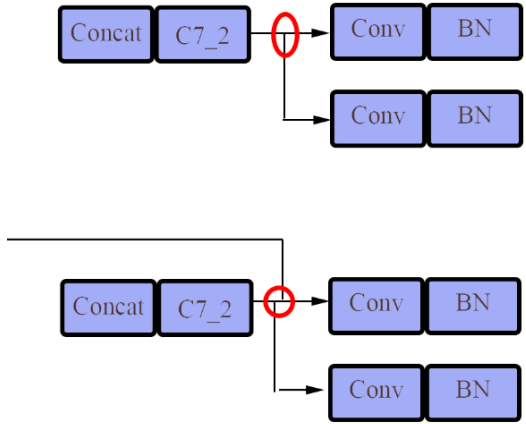

.

**Fig. 10.** Where the Self-attention added.

*B. Evaluation*

Since this is an instance segmentation task with only one class, False Negative (FN) is to identify the ocean in the background as the background. However, for satellite images containing plastic debris, the debris is only a small part of the image in most cases, so even if the debris detection error occurs, the accuracy rate is high (greater than 95%) due to the large area of the ocean background. Therefore, this part makes a more objective and accurate analysis of the performance of machine learning by calculating the precision and recall rate. In addition to precision and recall, which are commonly used to evaluate machine learning models, there is intersection over union (IoU), which is commonly used for object recognition.

Confusion matrices commonly used in machine learning are introduced to compute the results. It is divided into two dimensions, "actual" and "predicted", and divided into four categories: true positive (TP), True negative (TN), false positive (FP) and false negative (FN). Precision is the fraction of all samples that are predicted to be correct that are actually correct. In this work, the precision is the ratio when the pixel data in the prediction map and the pixels on the ground truth are all equal to one. Colloquially speaking, the precision representation is actually the ability of the model to correctly identify Marine debris as Marine debris.

For the BOX part of instance segmentation, self-attention model has the best precision score of 0.832, which means that this model has the most accurate positioning ability for Marine debris. For the MASK part, CBAM model has the best precision score of 0.787, which means that this model has the most accurate ability to describe the size and shape of Marine debris.

Recall is the proportion of all positive samples that are correctly identified as positive. In the context of this work, recall represents the proportion of Marine debris that is correctly identified. For the BOX aspect and the MASK aspect, coordinate attention model with C2f achieves the best Recall scores of 0.787 and 0.773, respectively, which indicates that the model can correctly identify more Marine debris.

REPLACE THIS LINE WITH YOUR MANUSCRIPT ID NUMBER (DOUBLE-CLICK HERE TO EDIT)

8Mean Average Precision (mAP) is a statistic to evaluate the performance of an object detection model, which represents the average of the mean accuracies across different classes. Average Precision (AP) is the area under the precision curve at different recall rates. Since this work has only one class (Marine debris), mAP is equivalent to AP. In terms of BOX, the best performance of mAP_0.5 is the dual-attention model with C2f, with 0.773. The best performance in MASK is coordinate attention model with C2f, and its mAP_0.5 is 0.743.

The F-Score is a measure of the performance of a classification model. It is the harmonic mean of precision and recall. A higher F-Score indicates that the model has a stronger ability to identify positive examples and a higher comprehensive ability. But the F-Score is also affected by a parameter β, which indicates the importance between precision and recall. When β is greater than 1, more attention is paid to recall. When β is less than 1, more attention is paid to the accuracy. When β is equal to 1, it is the commonly used F1-Score. In terms of F1 score, the CBAM model has the best score, 0.77 and 0.73, respectively, from the perspective of BOX or MASK. This represents its best performance as an instance segmentation model.

A common evaluation measure for object detection, neural network detectors, and semantic segmentation techniques is called Intersection of Union (IoU). Its concept can be used to measure the amount of overlap between the predicted region and the ground truth. When applied to instances with irregular borders, the IoU value is calculated in terms of pixels rather than the area of the box.

In theory, an IoU of 1 is the perfect model, but in practice an IoU of 0.5 is considered a good model. On the IoU evaluation, the CBAM model still achieves the best performance, achieving 0.62 for BOX and 0.58 for MASK.

*C. Statistical Comparison*

Since the resulting data is too large, two tables are plotted separately in this part, where Table 1 is the performance in terms of BOX and Table 2 is the performance in terms of MASK. Each model was trained more than twice and the best value was averaged.

As can be seen from Table 1, from the perspective of F1 score, almost all models with attention mechanism have better comprehensive performance than the original YOLOv7 or the updated YOLOv8. The C2f module in the architecture of YOLOv7 will increase the recall rate and reduce the precision, which means that more plastic waste can be accurately found in this task. However, while this can increase the average precision in some cases, it can lead to a decrease in overall performance in many cases. Therefore, in general, lightweight convolutional modules are not suitable for this architecture or Marine plastic detection task.

As can be seen from Table 2, the data of the Mask part is generally lower than that of the Box part, and YOLOv7 itself even has better performance than the later YOLOv8, which may be due to the fact that YOLO is more biased towards object detection. And the addition of attention mechanism does not bring significant improvement to the model. This may be due to the fact that implementing various attention mechanisms on the code refers more to object detection than semantic segmentation. Even so, CBAM model was able to achieve the best results.

Combining the two tables, it is clear that CBAM model has the best performance. Although the structure with coordinate attention layer is weaker than CBAM layer in performance, it has lower training cost, which is represented by half of the training time and lower occupancy than CBAM. Self-attention, which also has a lower training time, suffers the most despite its numerically worse performance, as will be shown in later sections. For the fit of different models and C2f lightweight, it is obvious that coordinate attention model and dual-attention model are more suitable for C2f modules with richer gradient information and more lightweight, while CBAM requires traditional coherent convolutional layers. Due to the poor numerical performance, it is difficult to judge whether self-attention model and C2f module have a good correlation.

TABLE I
STATISTICAL RESULT OF BOX

| Model | Precision | Recall | mAP_0.5 | F1-Score | IoU |
|---|---|---|---|---|---|
| YOLOv7 | 0.695 | 0.635 | 0.674 | 0.66 | 0.50 |
| YOLOv8 | 0.756 | 0.656 | 0.746 | 0.70 | 0.54 |
| Coordinate Attention model | 0.750 | 0.672 | 0.765 | 0.71 | 0.55 |
| Coordinate Attention model with C2f | 0.622 | 0.787 | 0.762 | 0.69 | 0.53 |
| CBAM model | 0.821 | 0.721 | 0.756 | 0.77 | 0.62 |
| CBAM model with C2f | 0.665 | 0.717 | 0.718 | 0.69 | 0.53 |
| Self-attention model | 0.832 | 0.541 | 0.675 | 0.66 | 0.49 |
| Self-attention model with C2f | 0.693 | 0.557 | 0.616 | 0.62 | 0.45 |
| Dual-attention model | 0.795 | 0.639 | 0.751 | 0.71 | 0.55 |
| Dual-attention model with C2f | 0.706 | 0.721 | 0.773 | 0.71 | 0.56 |

TABLE II
STATISTICAL RESULT OF MASK

| Model | Precision | Recall | mAP_0.5 | F1-Score | IoU |
|---|---|---|---|---|---|
| YOLOv7 | 0.787 | 0.609 | 0.696 | 0.69 | 0.52 |
| YOLOv8 | 0.686 | 0.639 | 0.675 | 0.66 | 0.49 |
| Coordinate Attention model | 0.737 | 0.639 | 0.671 | 0.68 | 0.52 |
| Coordinate Attention model with C2f | 0.636 | 0.773 | 0.743 | 0.70 | 0.53 |
| CBAM model | 0.787 | 0.689 | 0.716 | 0.73 | 0.58 |
| CBAM model with C2f | 0.679 | 0.721 | 0.737 | 0.70 | 0.54 |
| Self-attention model | 0.710 | 0.459 | 0.525 | 0.56 | 0.39 |
| Self-attention model with C2f | 0.663 | 0.475 | 0.553 | 0.55 | 0.38 |
| Dual-attention model | 0.692 | 0.541 | 0.593 | 0.61 | 0.43 |
| Dual-attention model with C2f | 0.773 | 0.508 | 0.578 | 0.61 | 0.44 |

*D. Images Comparison*

The data does not mean all things because manual labeling is subject to large errors. Figure 11 is a comparison of the output of multiple models, which shows that these models generally have good predictive ability and have little deviation from each other. The images of CBAM model use different form of output due to the interruption by Google Colab. It can be seen that almost all models predict the exact location of the Marine debris, and some of them have successfully detected small parts missed in the label. This led to all test images being reaudited to re-evaluate the model accuracy from an image perspective. The self-attention model obviously produces a lot of repeated judgments, however this allows it to get the most complete segmentation results in some scenes, as shown in Figure 12.

Figure 13 shows the Mask difference between Self-attention model and CBAM model for another plastic fragment that is long and close to the border. The CBAM model losses a part of debris next to the border of image while self-attention model correctly finds it.

The dual-attention model combining CBAM and Self-attention inherits some of the advantages of Self-attention, but is more conservative. Self-attention model detects a piece of unlabeled debris but it also identifies large swaths of the ocean as part of the debris. The error of this part is reduced through dual-attention model, but there is still a part of the tail that is difficult to identify whether it is Marine debris (Figure 14).

In any case, by comparing the output images, the performance and practical significance of the model can be more accurately evaluated. The model of self-attention has a lower score when calculating the result due to the phenomenon of duplicate detection and detecting the background as plastic debris. However, according to the image, a large proportion of these cases of classifying the background as debris are identifying the missing parts of the label, so the actual performance of self-attention model should be better than the performance of data evaluation.

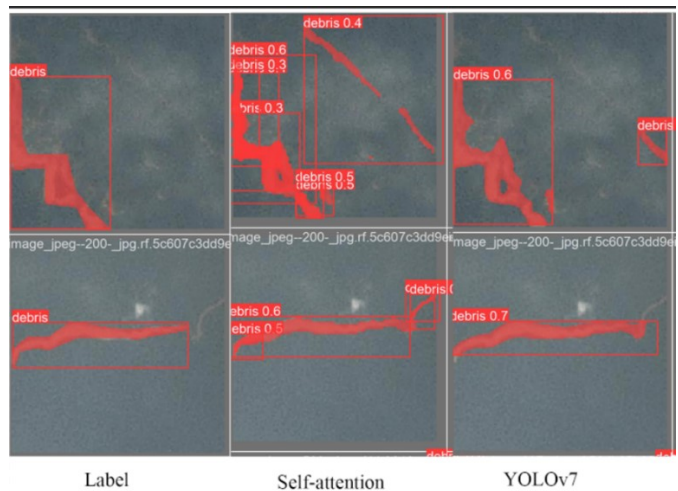

**Fig. 12.** A comparison between label, self-attention model and YOLOv7 model. Self-attention model has a complete detection of Marine debris.

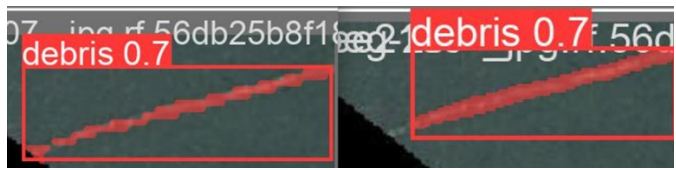

**Fig. 13.** Detailed comparison between self-attention model and CBAM model in some scenes. Left is self-attention model and right is CBAM model.

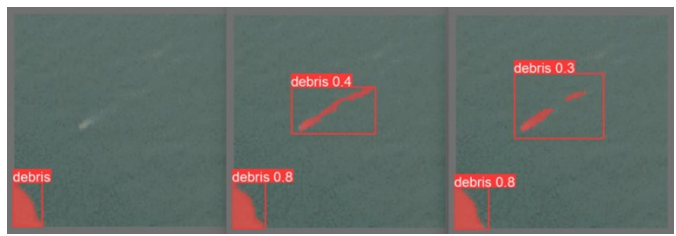

**Fig. 14.** Comparison between label, self-attention model and dual-attention model. The left one is label, the middle one is a self-attention model, and the right one is a dual-attention model.

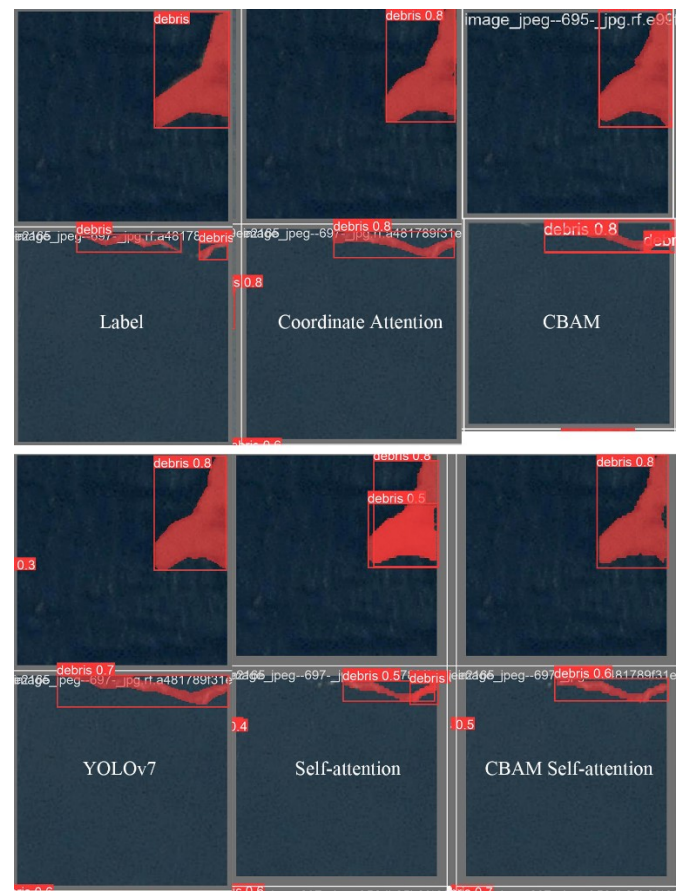

**Fig. 13.** A general comparison of improved models.

*E. Result Analysis*

Through the comparison of eight models including C2f, it can be found that C2f seems to have little effect on the comprehensive performance of the model. Its performance is mainly reflected in the single item value. Therefore, the model with C2f module and attention mechanism is counted as the same model as the model with only attention mechanism. For example, dual-attention model and dual-attention model with C2f both count as dual-attention model. Therefore, although the gap between them seems not to be very large from the numerical point of view (especially in terms of Mask), it can be seen that the CBAM model has the best comprehensive performance, followed by the coordinate attention model, combined with the detail comparison on the image. This is followed by the dual-attention model, then the YOLOv7 instance segmentation itself, and the worst is the self-attention model.

It can be seen that the performance of the dual-attention model with CBAM decreases to a level similar to that of the coordinate attention model due to the addition of the self-attention block. In terms of comprehensive performance and single attribute (precision and recall), the dual-attention model is more balanced than the coordinate attention model. However, dual-attention model is more biased than coordinate attention model when considering Mask and Box simultaneously. The training cost (training time) of the dual-attention model is almost the same as that of the coordinate attention model, which is about 25% longer than that of the self-attention model. Therefore, in cases where it is difficult to afford the higher training cost of CBAM model, the coordinate attention model can be prioritized.

Unlike YOLOv8, the model of YOLOv7 instance segmentation can obviously be used for plastic waste detection tasks in satellite images with only the sea surface, however, its comprehensive performance is indeed lower than several improved models proposed in this work. The Self-attention model is affected by a series of restrictions, which has great limitations for its evaluation, so its data is not reliable and has less practical significance, and more work is still needed to test it.

V. DISCUSSION

This research seeks to integrate YOLOv7 instance segmentation with diverse attention mechanisms towards enhancing the detection of marine debris in satellite imagery. The goal is to establish a reliable, efficient, and extensible real-time monitoring system for identifying marine plastic waste. Overall, the nine models tested in this work, including YOLOv7, demonstrate satisfactory ability to identify marine debris. While specific metrics like precision might seem subpar, factors like recall, F1-score, and visual analysis indicate promising results. Among all the modified versions evaluated, only the self-attention variant performed worse than YOLOv7, while others surpassed or matched its accuracy. The CBAM model yielded top outcomes. Coordinate Attention and Dual Attention also exhibited competitive yet affordable capabilities. Nevertheless, keep in mind that these observations do not necessarily imply the inferiority of self-attention over YOLOv7; additional experiments with richer data sets would be necessary to draw definitive conclusions.

The dataset was insufficient and biased towards specific situations. Satellite images with a resolution of 3m/pxl exclusively featuring sea surfaces remain hard to find online. A large portion of these datasets consists of images of land areas such as beaches. Websites like USGS Earth Explorer and Sentinel Open Access Hub display no data or only blurred images when zooming in on the ocean section. Consequently, the dataset remains quite scarce, and homogeneous, and fails to cover various weather settings, further reducing the chances of the attention mechanism performing optimally. Small sample sizes impede the model's capacity to discern among diverse lightning and cloud densities.

Furthermore, implementing and tweaking attention mechanisms is not always suitable for detecting maritime waste via satellite imagery. Just as YOLOv8, which performed better on COCO dataset, did not perform as well as YOLOv7 in this work. Although these attention mechanisms have been experimented with and their high performance has been verified, their application scenarios are still limited.

Two primary paths could branch off from this work: one focused on improving applications in the marine domain, and another centered on advancing deep learning techniques.

In terms of marine usage, there are limitations due to insufficient dataset diversity, which affects attention mechanism performance. To develop a robust and versatile marine debris recognition system, access to high-quality images from various environments, light conditions, and geographical regions would be essential. Additionally, using spectrograms instead of RGB images, which capture more detailed information, could enhance the effectiveness of the solution. Moreover, integrating oceanographic knowledge, particularly regarding sea currents, would help forecast the movement of marine waste, making the platform even more valuable. Finally, this work uses YOLO as a base to expand the possibilities of real-time detection. This real-time doesn't exactly mean streaming media, but rather, depending on how the satellite collects the data, it can mean quick preliminary processing and classification of the images. This is also a feature worth exploring.

As for deep learning improvements, the viability of self-attention mechanisms in plastic debris identification was proven during this investigation. Exploring novel ways of applying self-attention without relying on manually labeled data could streamline the process significantly. Unsupervised learning combined with self-attention holds enormous promise for cutting down human effort while boosting efficacy. Conducting extensive experimental evaluations of self-attention variants will pave the way forward toward developing advanced AI tools tailored specifically for addressing ecological issues, notably those involving the oceans.

VI. CONCLUSION

In conclusion, our work demonstrates the potential of attention mechanisms in improving the accuracy and efficiency of marine debris detection in satellite imagery. Our experiments show that CBAM is the most suitable attentional model for this task, but the bottleneck transformer may offer better practical performance in certain scenarios. We also identify several areas for further research, including the need for more diverse datasets and the potential benefits of using spectrograms and integrating oceanographic knowledge. Overall, we believe that our findings can contribute to the development of more robust and versatile marine debris recognition systems, which can ultimately aid in the global effort to reduce plastic pollution in our oceans.

ACKNOWLEDGMENT

This work was supported in part by the Engineering and Physical Sciences Research Council (EPSRC) Grant EP/P009727/2 and in part by the Leverhulme Trust Grant RF-2019-492.